\documentclass{article} 
\usepackage{iclr2015,times}
\usepackage{hyperref}
\usepackage{url}

\usepackage{latexsym}           %
\usepackage{times}              %
\usepackage{amsfonts}           %
\usepackage{amsmath}            %
\usepackage{graphicx}           %
\usepackage{caption}
\usepackage{multirow}
\usepackage{booktabs}
\usepackage{subcaption}
\usepackage{empheq}
\usepackage{float}
\usepackage[utf8]{inputenc}     %

\newcommand{\field}[1]{\mathbb{#1}}                                            %
\newcommand{\Hset}{\mathcal{Q}}                                                %
\newcommand{\Mset}{\mathcal{C}}                                                %
\newcommand{\Dset}{\mathcal{D}}                                                %
\title{Tailoring Word Embeddings For Bilexical Predictions: An Experimental Comparison}
\author{
Pranava Swaroop Madhyastha\\
Universitat Polit\`{e}cnica de Catalunya\\
Campus Nord UPC, 08034 Barcelona \\
\texttt{pranava@cs.upc.edu} \\
\And
Xavier Carreras \\
Xerox Research Centre Europe\\
38240 Meylan, France\\
\texttt{xavier.carreras@xrce.xerox.com} \\
\And Ariadna Quattoni\\
Xerox Research Centre Europe\\
38240 Meylan, France \\
\texttt{ariadna.quattoni@xrce.xerox.com}
}
\iclrfinalcopy 
\begin{document}
\maketitle

\begin{abstract}
We investigate the problem of inducing word embeddings that are
tailored for a particular bilexical relation. Our learning algorithm
takes an existing lexical vector space and compresses it such that the
resulting word embeddings are good predictors for a target bilexical
relation. In experiments we show that task-specific embeddings can
benefit both the quality and efficiency in lexical prediction tasks.

\end{abstract}

\section{Introduction}

There has been a large body of work that focuses on
learning word representations, either in the form of word clusters
\citep{Brown92} or vectors \citep{sahlgrenthesis,Turney+Pantel-2010,mikolov2013,
  glove,baroni-dinu-kruszewski:2014:P14-1,bansal2014tailoring} and these have proven
useful in many NLP applications
\citep{Koo+08,turian-ratinov-bengio:2010:ACL}.

An ideal lexical representation should compress the space of lexical
words while retaining the essential properties of words in order to
make predictions that correctly generalize across words. The typical
approach is to first induce a lexical representation in a
task-agnostic setting and then use it in different tasks as
features. A different approach is to learn a lexical representation
tailored for a certain task. In this work we explore the second
approach, and employ the formulation by \cite{coling-2014} to induce
task-specific word embeddings. This method departs from a given
lexical vector space, and compresses it such that the resulting word
embeddings are good predictors for a given lexical relation.

Specifically we learn functions that compute compatibility scores
between pairs of lexical items under some linguistic relation.  In our
work, we refer to these functions as bilexical operators. As an
instance of this problem, consider learning a model that predicts the
probability that an adjective modifies a noun in a sentence. In this
case, we would like the bilexical operator to capture the fact that
some adjectives are more compatible with some nouns than others.

Given the complexity of lexical relations, one expects that the
properties of words that are relevant for some relation are different
for another relation. This might affect the quality of an embedding,
both in terms of its predictive power and the compression it
obtains. If we employ a task-agnostic low-dimensional embedding, will
it retain all important lexical properties for any relation? And,
given a fixed relation, can we further compress an existing word
representation? In this work we present experiments along these lines
that confirm that task-specific embeddings can benefit both the
quality and the efficiency of lexicalized predictive models.

\section{Formulation}
\newcommand{\VOC}{\mathcal{V}}

Let $\VOC$ be a vocabulary, and let $x \in \VOC$ denote a word.  We
are interested in modeling a target bilexical relation, that is, a relation
between pairs of words without context. For example, in a
noun-adjective relation we model what nouns can be assigned to
what adjectives. We will denote as $\Hset \subseteq \VOC$ the set of query
words, or words that appear in the left side of the bilexical
relation. And we will use $\Mset \subseteq \VOC$ to denote
\emph{candidate} words, appearing in the right side of the relation.

In this paper we experiment with the log-linear models by
\cite{coling-2014} that given a query word $q$ compute a conditional
distribution over candidate words $c$. The models take the following
form:
\begin{equation}
\Pr( c \mid q ; W ) = \frac{ \exp \{ \phi(q)^\top W \phi(c) \}}{ \sum_{c'} \exp \{ \phi(q)^\top W \phi(c') \}}
\end{equation}
where $\phi: \VOC \to \field{R}^n$ is a distributional representation of
words, and $W \in \field{R}^{n \times n}$ is a bilinear form. 

The learning problem is to obtain $\phi$ and $W$ from data, and we
approach it in a semi-supervised fashion. There exist many approaches
to learn $\phi$ from unlabeled data, and in this paper we experiment
with two approaches: (a) a simple distributional approach where we
represent words with a bag-of-words of contextual words; and (b) the
skip-gram model by \cite{mikolov2013}. To learn $W$ we assume access to
labeled data in the form pairs of compatible examples, i.e.
$\Dset=\{(q,c)^1, \ldots, (q,c)^l\}$, where $q \in \Hset$ and $c \in
\Mset$. The goal is to be able to predict query-candidate pairs that
are unseen during training. Recall that we model relations between
words without context. Thus the lexical representation $\phi$ is
essential to generalize to pairs involving unseen words.

With $\phi$ fixed, we learn $W$ by minimizing the negative
log-likelihood of the labeled data using a regularized objective,
$L(W) = -\sum_{ (q,c) \in D} \log \Pr(c \mid q; W) + \tau \rho(W)$,
where $\rho(W)$ is a regularization penalty and $\tau$ is a constant
controlling the trade-off.

We are interested in regularizers that induce low-rank parameters $W$,
since they lead to task-specific embeddings.  Assume that $W$ has rank
$k$, such that $W = U V^\top$ with $U, V \in \field{R}^{n \times
  k}$. If we consider the product $\phi(q)^\top U V^\top \phi(c)$, we
can now interpret $\phi^\top U$ as a $k$-dimensional embedding of
query words, and $\phi(c)^\top V$ as a $k$-dimensional embedding of
candidate words. Thus, if we obtain a low-rank $W$ that is highly
predictive, we can interpret $U$ and $V$ as task-specific compressions
of the original embedding $\phi$ tailored for the target bilexical
relation, from $n$ to $k$ dimensions.

Since minimizing the rank of a matrix is hard, we employ a convex
relaxation based on the nuclear norm of the matrix $\ell_\star$ (that is, the $\ell_1$ norm of the singular values, see \citet{Srebro05}). In
our experiments we compare the low-rank approach to $\ell_1$ and
$\ell_2$ regularization penalties, which are common in linear
prediction tasks. For all settings we use the \emph{forward-backward
  splitting (FOBOS)} optimization algorithm by
\cite{duchi2009efficient}.

We note that if we set $W$ to be the identity matrix our model scores
are inner products between the query-candidate embeddings, a common
approach to evaluate semantic similarity in unsupervised
distributional approaches. In general, we can compute a
low-dimensional projection of $\phi$ down to $k$ dimensions, using
SVD, and perform the inner product in the projected space. We refer to
this as the unsupervised approach, since the projected embeddings do
not use the labeled dataset specifying the target relation.

\section{Experiments with Syntactic Relations}
\begin{table}[h]
    \centering
    \resizebox{.6\textwidth}{!}{%
        \begin{tabular}{lcclcccc}
            \toprule
            & & \multicolumn{5}{ c }{$\ell_\star$} \\ 
            \cmidrule(r){4-6}
            {\it Rel} & {\it Type} &  {\it UNS} & \multicolumn{1}{c}{\it best $k$}  & $k=5$ & $k=10$ &  $\ell_2$ & $\ell_1$ \\
            \midrule
                Adj-Noun & BoW & 85.12 & 85.99 \ \ (80) & 83.99 & 84.74 & 85.96 & 85.63 \\
                     & SKG & 73.61 & 91.40 \ (300) & 83.70 & 86.27 & 91.22 & 90.72 \\
            \midrule
                Obj-Verb & BoW & 63.85 & 78.11 \ (200) & 73.17 & 73.64 & 74.08 & 73.95 \\
                     & SKG &  64.15 & 79.98 \ \ (50) & 75.45 & 78.37 & 80.30 & 79.89 \\
            \midrule
                Subj-Verb & BoW & 58.20 & 78.13 \ \ \ (2) & 71.71 & 71.73 & 78.07 & 77.97 \\
                      & SKG & 49.65 & 59.28 \ \ (90) & 53.31 & 53.32 & 58.24 & 58.67 \\
            \midrule
                Noun-Adj & BoW & 78.09 & 78.11 \ \ (70) & 77.48 & 77.85 & 78.48 & 78.85 \\
                         & SKG  &  49.65 & 59.28 \ \ (50) & 56.42 & 57.19 &  58.24 & 58.67 \\
            \midrule
                Verb-Obj & BoW & 66.46 & 73.90 \ \ (40) & 73.70 & 73.88 & 73.30 & 73.48 \\
                         & SKG     &  64.15 & 79.99 \ \ (30) & 77.05 & 78.60 & 80.29 & 79.89 \\
            \midrule
                Verb-Subj & BoW  & 49.32 & 71.97 \ \ (30) & 71.71 & 71.23 & 72.85 & 71.95 \\
                          & SKG & 32.34 & 53.75 \ \ \  (2) & 53.32 & 53.32 & 53.47 & 53.68 \\
            \midrule
        \end{tabular}
    }
    \caption{\small{Pairwise accuracies for the six relations using
        the unsupervised, $\ell_\star$, $\ell_2$ and $\ell_1$ models,
        using either a distributional bag-of-words representation
        (BoW) or the skip-gram embeddings (SKG) as initial
        representation. For $\ell_\star$ we show results for the rank
        that gives best accuracy (with the optimal rank in
        parenthesis), as well as for ranks $k=5$ and $10$.  }}
    \label{tablenava}
\end{table}
\begin{figure}[p!]
    \def\wi{0.48}
    \def\tmps{0.85}
    \hskip 1.2cm
    \begin{subfigure}[t]{\wi\textwidth}
        \includegraphics[scale=0.9]{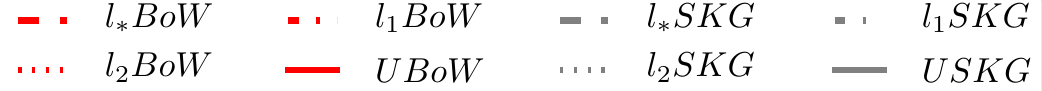}
    \end{subfigure}

    \begin{subfigure}[t]{\wi\textwidth}
        \caption{Adjective-Noun}
        \includegraphics[scale=\tmps]{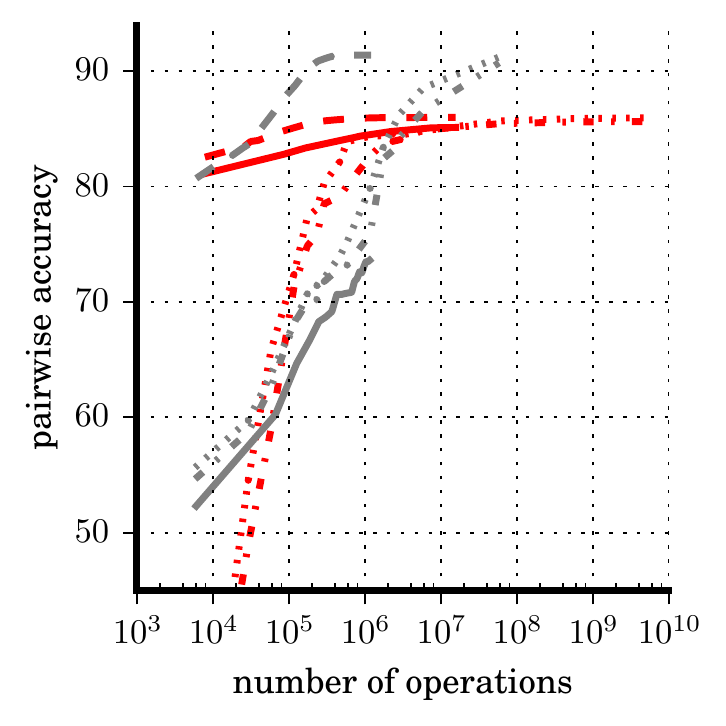}
    \end{subfigure}
    \begin{subfigure}[t]{\wi\textwidth}
        \caption{\small Noun-Adjective}
        \includegraphics[scale=\tmps]{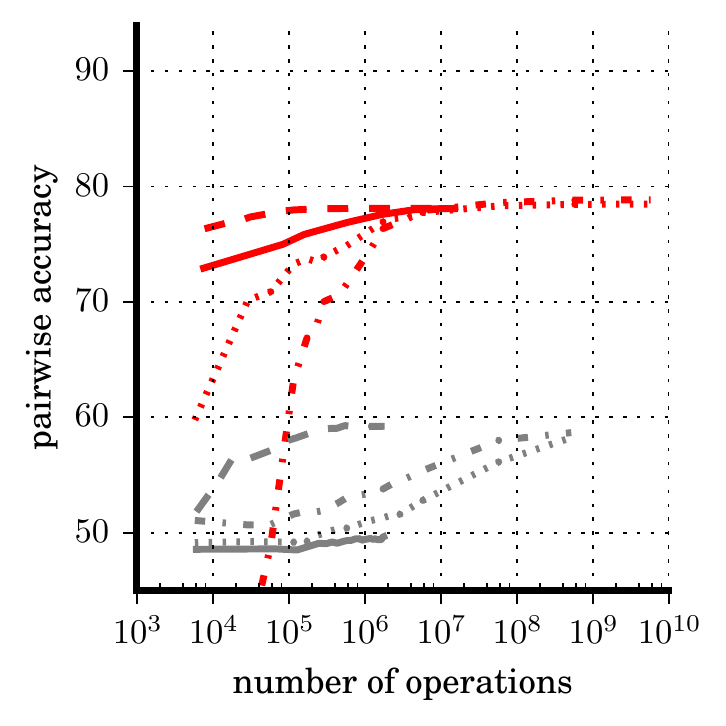}
    \end{subfigure}
    \begin{subfigure}[t]{\wi\textwidth}
        \caption{\small Verb-Object}
        \includegraphics[scale=\tmps]{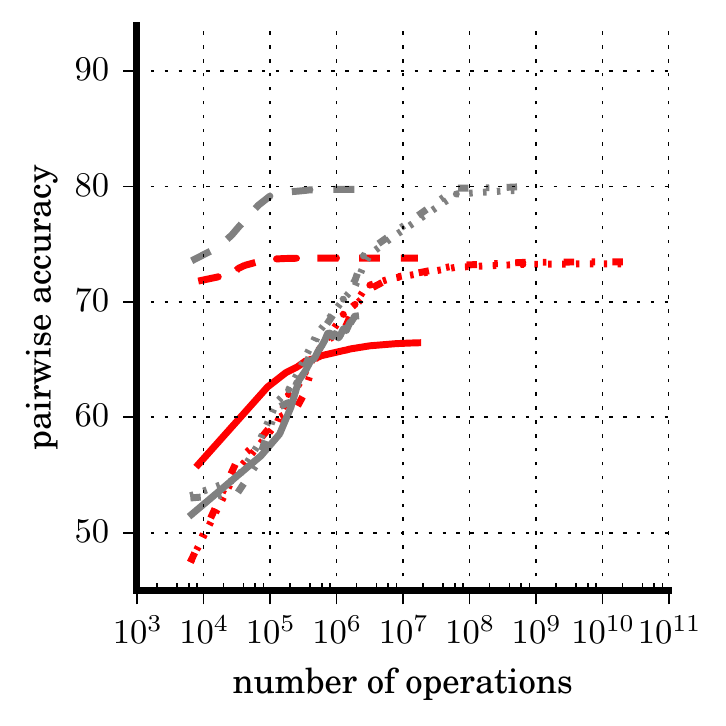}
    \end{subfigure}
    \begin{subfigure}[t]{\wi\textwidth}
        \caption{\small Object-Verb}
        \includegraphics[scale=\tmps]{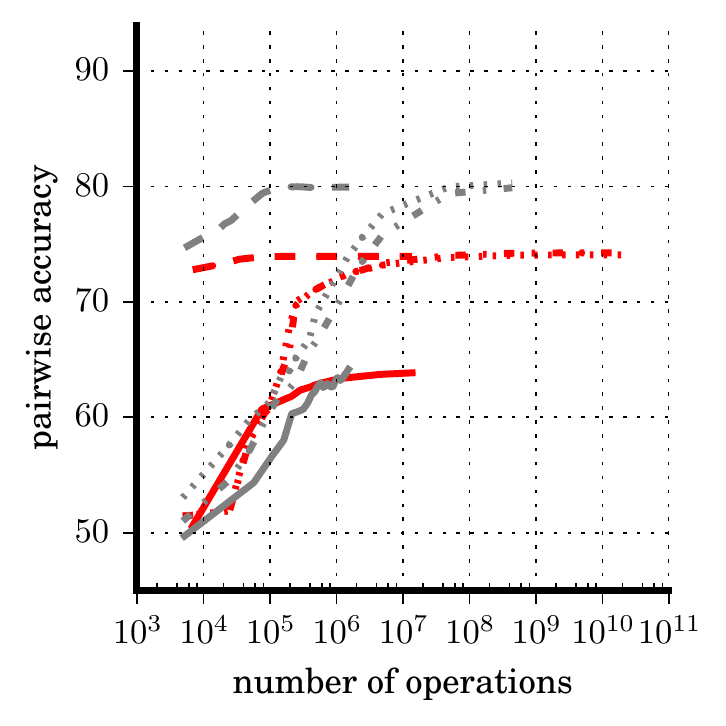}
    \end{subfigure}
    \begin{subfigure}[t]{\wi\textwidth}
        \caption{\small Subject-Verb}
        \includegraphics[scale=\tmps]{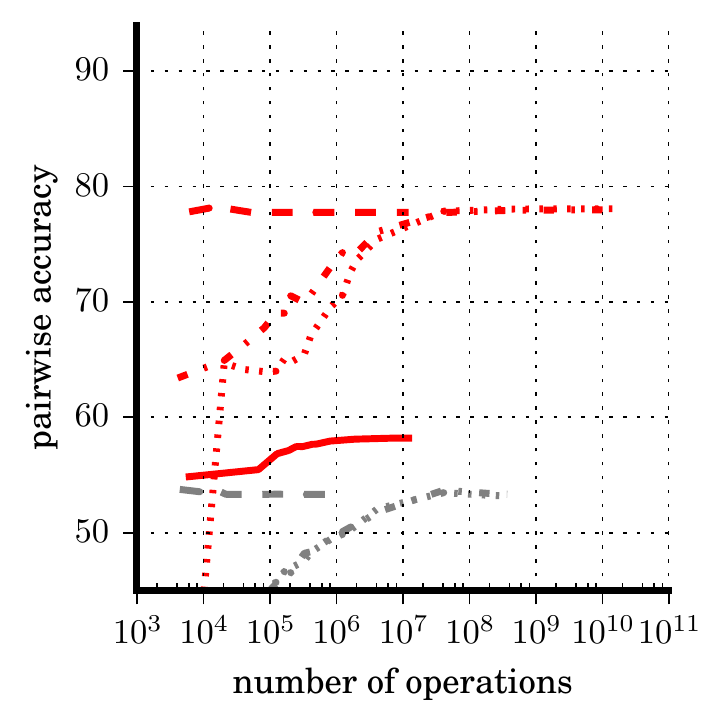}
    \end{subfigure}
    \begin{subfigure}[t]{\wi\textwidth}
        \caption{\small Verb-Subject}
        \includegraphics[scale=\tmps]{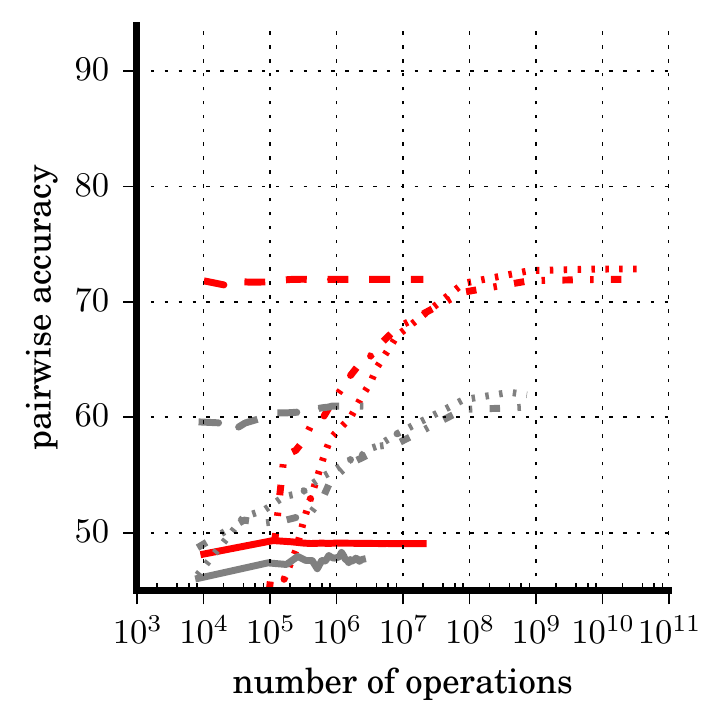}
    \end{subfigure}

    \caption{\small{Pairwise accuracy v/s no. of double operations to
        compute the distribution over candidate words for a query
        word.  Plots are for noun-adjective, verb-object and
        verb-subject relations, in both directions. The red curves use
        distributional representation based on bag-of-words (BoW) and
        the grey curves use the embeddings of the skip-gram model (SKG).}}
\label{fig1}
\end{figure}

We conducted a set of experiments to test the performance of the
learning algorithm with respect to the initial lexical representation
$\phi$, for different configurations of the representation and the
learner. We experiment with six bilexical syntactic relations using
the Penn Treebank corpus \citep{penntreebank}, following the
experimental setting by \cite{coling-2014}. For a relation between
queries and candidate words, such as noun-adjective, we partition the
query words into train, development and test queries, thus test
pairs are always unseen pairs.

To report performance, we measure pairwise accuracy with respect to
the efficiency of the model in terms of number of active
parameters. To measure the efficiency of a model we consider the
number of double operations that are needed to compute, given a query
word, the scores for all candidates in the vocabulary. See
\citep{coling-2014} for details.

We experiment with two types of initial representations $\phi$. The
first is a simple high-dimensional distributional representation based
on contextual bag-of-words (BoW): each word is represented by the bag
of words that appear in contextual windows. In our experiments these
were sparse 2,000-dimensional vectors.  The second representation are the
low-dimensional skip-gram embeddings (SKG) by \cite{mikolov2013}, where
we used 300 dimensions. In both cases we induced such representations
using the BLIPP corpus \citep{bllip} and using a context window of
size 10 for both. Thus the main difference is that the bag-of-words is
an uncompressed representation, while the skip-gram embeddings are a
neural-net-style compression of the same contextual windows.

As for the bilexical model, we test it under three regularization
schemes, namely $\ell_2$, $\ell_1$, and $\ell_\star$. For the first
two, the efficiency of computing predictions is a function of the
non-zero entries in $W$, while for the latter it is the rank $k$ of
$W$, which defines the dimension of the task-specific embeddings. We
also test a baseline unsupervised approach (UNS).

\section{Results and Discussion}

Figure~\ref{fig1} shows the performance of models for noun-adjective,
verb-object and verb-subject relations (in both directions).  
In line with the results by \cite{coling-2014} we observe that
the supervised approach in all cases outperforms the unsupervised case,
and that the nuclear norm scheme provides the best performance in
terms of accuracy and speed: other regularizers can obtain similar
accuracies, but low-rank constraints during learning favor very-low
dimensional embeddings that are highly predictive.

In terms of starting with bag-of-words vectors or skip-gram
embeddings, in three relations the former is clearly better, while in
the other three relations the latter is clearly better. We conclude
that task-agnostic embeddings do identify useful relevant properties
of words, but at the same time not all necessary properties are
retained. In all cases, the nuclear norm regularizer successfully
compresses the initial representation, even for the embeddings which
are already low-dimensional.

Table~\ref{tablenava} presents the best result for each relation,
initial representation and regularization scheme. Plus, for the
$\ell_\star$ regularizer we present results at three different ranks,
namely 5, 10 or the rank that obtains the best result for each
relation. These highly compressed embeddings perform nearly as good as
the best performing model for each relation.

\begin{table}[t]
    \centering
    \resizebox{\textwidth}{!}{%
    \begin{tabular}{||l|l|l||}
        \hline
    \multicolumn{1}{|c}{\bfseries Query} & \multicolumn{1}{|c}{\bfseries noun-adjective} & \multicolumn{1}{|c|}{\bfseries object-verb}  \\ \hline
        \hline
        city & province, area, island, township, freeways  & residents, towns, marchers, streets, mayor \\
        \hline
        securities & bonds, mortgage, issuers, debt, loans  & bonds, memberships, equities, certificates, syndicate \\
        \hline
        board & committee directors, commission, nominees, refusal & slate, membership, committee, appointment, stockholder \\
        \hline
        debt & loan, loans, debts, financing, mortgage & reinvestment, indebtedness, expenditures, outlay, repayment \\
        \hline
        law & laws, constitution, code, legislation, immigration &  laws, ordinance, decree, statutes, state \\
        \hline
        director & assistant, editor, treasurer, postmaster, chairman & firm, consultant, president, manager, leader \\
        \hline
    \end{tabular}
    }
    \caption{Example query words and 5 highest-ranked candidate
      words for two different bilexical relations: noun-adjective and
      object-verb.}
    \label{tableembed}
\end{table}

Table~\ref{tableembed} shows a set of query nouns, and two sets of
neighbor query nouns, using the embeddings for two different relations
to compute the two sets. We can see that, by changing the target
relation, the set of close words changes. This suggests that words have
a wide range of different behaviors, and different relations might
exploit lexical properties that are specific to the relation.

\section{Conclusion}

We have presented a set of experiments where we compute word
embeddings specific to target linguistic relations.
We observe that low-rank penalties favor embeddings that are good both in terms of
predictive accuracy and efficiency.
For example, in certain cases, models using very low-dimensional
embeddings perform nearly as good as the best models.

In certain tasks, we have shown that we can refine low-dimensional
skip-gram embeddings, making them more compressed while retaining
their predictive properties. In other tasks, we have shown that our
method can improve over skip-gram models when starting from
uncompressed distributional representations. This suggests that
skip-gram embeddings do not retain all the necessary information of
the original words. This motivates future research that aims at
general-purpose embeddings that do retain all necessary properties,
and can be further compressed in light of specific lexical relations.

\section*{Acknowledgements}
We thank the reviewers for their helpful comments. 
This work has been partially funded by the Spanish Government through
the SKATER project (TIN2012-38584-C06-01).

\bibliographystyle{iclr2015}
\bibliography{ref}

\end{document}